\documentclass[letterpaper, 10 pt, conference]{ieeeconf}  % Comment this line out if you need a4paper

\IEEEoverridecommandlockouts                              % This command is only needed if 
\usepackage{graphicx} % for \includegraphics
\usepackage[export]{adjustbox} % for \adjincludegraphics
\usepackage[skip=1pt,labelformat=empty]{subcaption} % for subfigure caption skip setting, and remove label format
\usepackage[cmex10]{amsmath} % assumes amsmath package installed
\usepackage[short]{optidef} % for argmini environment
\usepackage[ruled,linesnumbered]{algorithm2e} % for pseudocode
\usepackage{caption} % for figure caption skip setting
\usepackage{bm} % for \bm
\usepackage{epsfig} % for postscript graphics files
\usepackage{times} % assumes new font selection scheme installed
\usepackage{color} % for \color
\usepackage{amssymb}  % assumes amsmath package installed
\usepackage{setspace}

\newcommand{\mytilde}{\raise.17ex\hbox{$\scriptstyle\mathtt{‌​\sim}$}}
\usepackage{xparse,soul}
\usepackage{multirow}
\usepackage{cite}
\usepackage[font=small,labelfont=bf]{caption}

\usepackage[utf8]{inputenc}
\usepackage{mathtools}

\title{\LARGE \bf
APS-LSTM: Exploiting Multi-Periodicity and Diverse Spatial Dependencies for Flood Forecasting
}

\author{Jun Feng$^{1,2}$\textsuperscript{\textdagger}, 
Xueyi Liu$^{1,2}$,
Jiamin Lu$^{1,2}$,
Pingping Shao$^{1,2}$% <-this % stops a space
\thanks{The work is supported in part by the National Key R\&D Program of China (Grant No. 2021YFB3900601).}%
\thanks{$^{1}$Key Laboratory of Water Big Data Technology of Ministry of Water Resources, Hohai University, 211100 Nanjing, China.}
\thanks{$^{2}$College of Computer and Software, Hohai University, 211100 Nanjing, China.}
\thanks{\textsuperscript{\textdagger}Corresponding author, email: fengjun@hhu.edu.cn.}
}

\begin{document}

\maketitle
\thispagestyle{empty}
\pagestyle{empty}

%%%%%%%%%%%%%%%%%%%%%%%%%%%%%%%%%%%%%%%%%%%%%%%%%%%%%%%%%%%%%%%%%%%%%%%%%%%%%%%%

\begin{abstract}
Accurate flood prediction is crucial for disaster prevention and mitigation. Hydrological data exhibit highly nonlinear temporal patterns and encompass complex spatial relationships between rainfall and flow. Existing flood prediction models struggle to capture these intricate temporal features and spatial dependencies. This paper presents an adaptive periodic and spatial self-attention method based on LSTM (APS-LSTM) to address these challenges. The APS-LSTM learns temporal features from a multi-periodicity perspective and captures diverse spatial dependencies from different period divisions. The APS-LSTM consists of three main stages, (i) Multi-Period Division, that utilizes Fast Fourier Transform (FFT) to divide various periodic patterns; (ii) Spatio-Temporal Information Extraction, that performs periodic and spatial self-attention focusing on intra- and inter-periodic temporal patterns and spatial dependencies; (iii) Adaptive Aggregation, that relies on amplitude strength to aggregate the computational results from each periodic division. The abundant experiments on two real-world datasets demonstrate the superiority of APS-LSTM. The code is available: https://github.com/oopcmd/APS-LSTM.
\end{abstract}

%%%%%%%%%%%%%%%%%%%%%%%%%%%%%%%%%%%%%%%%%%%%%%%%%%%%%%%%%%%%%%%%%%%%%%%%%%%%%%%%
\section{INTRODUCTION}
Floods are one of the most destructive natural disasters, which can cause serious damage and loss to human life and production, affect the development and stability of social economy, and even endanger human life safety~\cite{he2018analysis}. Therefore, how to accurately predict the occurrence time, development trend and impact range of floods, and take effective prevention and response measures according to the prediction results, is an important and urgent problem we face. Accurate flood forecasting has become a key task in disaster prevention and mitigation strategies.

Data-driven flood forecasting can essentially be reduced to a time series forecasting problem. With the rapid development of deep learning, many time series forecasting models~\cite{lstm,ts-method1,ts-method2} have emerged, which model data from the temporal dimension and learn the interrelationships between different time points to predict future time series. Additionally, many recent models rely on static periodic patterns~\cite{fixperiod1,fixperiod2} and seasonal trend decomposition~\cite{arima,Dlinear} for time series forecasting. These approaches have demonstrated significant improvements in prediction accuracy when viewed from a periodic perspective. However, they apparently fail to realize the complex and overlapping nature of periodic features in real-world data. Specifically, hydrological data exhibits hidden and intricate multi-periodicity due to various natural factors, such as rainfall, terrain, soil, vegetation, etc. Methods based on fixed periods and trend decomposition are inadequate for accurately handling the multiple interacting periods present in hydrological data.

In addition, spatial correlations also play a crucial role in accurate flood forecasting for small and medium-sized watersheds. Recently, many spatio-temporal prediction methods have leveraged Graph Convolutional Networks (GCNs)~\cite{gcn,gcn1,gcn2}, learnable parameters~\cite{st-id,st-mlp}, and attention mechanisms~\cite{astgcn,astgnn} to capture dynamic spatial information. While these methods show promise, they treat the entire input data as a single period, attempting to learn spatial information within this fixed period. In hydrological data, the spatial contribution of rainfall to flow is exceedingly complex and difficult to disentangle. The spatial correlations learned from a singular perspective clearly fall short in effectively representing the complex spatial structures present in the real-world.

\begin{figure}[h]
    \centering
    \includegraphics[width=\linewidth]{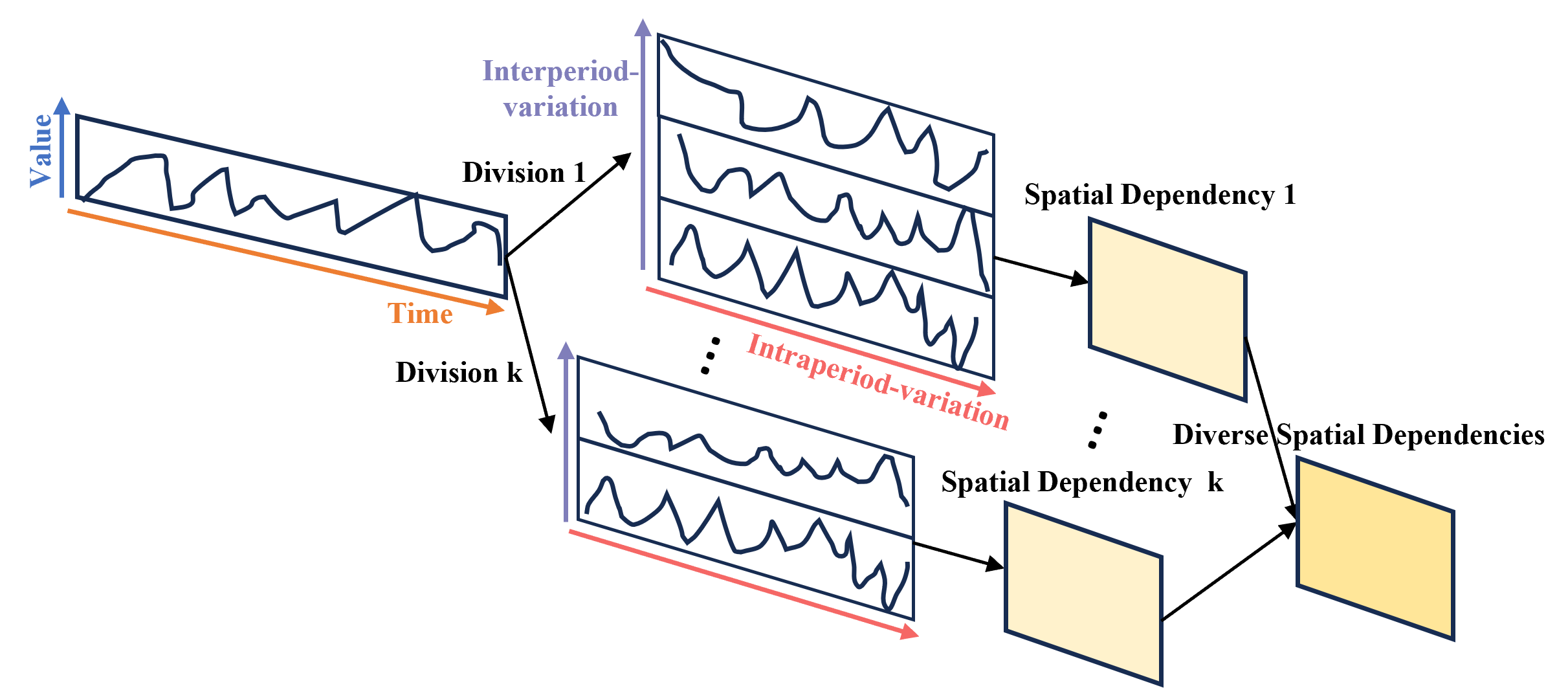}
    \caption{Illustration of Diverse Spatial Dependencies}
    \label{fig:dsd}
    \vspace{-0.4cm}
\end{figure}

To address the above challenges, inspired by TimesNet~\cite{timesnet}, we introduce the Adaptive Periodic and Spatial Self-Attention model based on LSTM (APS-LSTM). Fig.~\ref{fig:dsd} illustrates the diverse spatial dependencies, which are composed of spatial relationships learned under varying periodic divisions. The existing spatial information learning methods are merely a specific instance of diverse spatial dependencies. Specifically, APS-LSTM first uses Fast Fourier Transform (FFT) to convert time-domain data into frequency-domain, dividing distinct periods based on the most contributory frequencies. Within each division, periodic self-attention mechanisms are deployed to accentuate intra- and inter-periodic information. Advancing further, spatial self-attention mechanisms are utilized to excavate concealed dynamic spatial structures within each period division. Then, we perform adaptive aggregation according to the importance of each period and frequency, and obtain data containing multi-periodicity and diverse spatial dependencies. Leveraging the superior sequential modeling prowess of Long Short Term Memory (LSTM)~\cite{lstm}, the richly represented spatio-temporal data is effectively encoded, and subsequently decoded through a linear output layer. The principal contributions of this paper are as follows:
\begin{itemize}
    \item A novel model named APS-LSTM is proposed, which, for the first time, applies multi-periodicity and diverse spatial dependencies for flood forecasting.
    
    \item APS-LSTM leverages periodic and spatial self-attention mechanisms to effectively capture spatio-temporal features within and across periods under various divisions. By adaptively aggregating these features, it opens up new avenues for exploring more comprehensive and efficient spatio-temporal representations in hydrological data.

    \item Through the experiments on two real-world hydrological datasets, the empirical evidence indicates that the APS-LSTM surpasses the established benchmarks, achieving the best performance.
\end{itemize}

\section{RELATED WORK}
\subsection{Data-Driven Flood Prediction}
Earlier flood prediction methods generally used statistical approaches, e.g., support vector machine~\cite{svm}, but with the development of neural networks, data-driven flood prediction methods have been improved. STA-LSTM~\cite{sta-lstm} proposed by Ding et al. incorporates temporal and spatial attention into LSTM and is used for flood prediction. Sha et al.~\cite{gcnlstm} combined graph convolutional networks with LSTM while also introducing temporal and spatial attention and successfully used it for flood prediction. Unfortunately, the aforementioned flood prediction methods do not consider the complex multi-periodicity and diverse spatiality in hydrological data, which are the focus of this paper.

\subsection{Spatio-Temporal Sequence Prediction}
GNNs are widely used in temporal sequence prediction tasks with spatial information. Graph WaveNet~\cite{GWnet} employs GCN combined with dilated convolution and adaptive dependency matrix to capture spatio-temporal information in the traffic data. ASTGNN~\cite{astgnn} applies GCN to Transformer architecture, and learns temporal trend by time trend-aware self-attention. In recent years, many methods have focused their attention on the trend and periodicity of time series. Autoformer~\cite{Autoformer} extracts the seasonal trend from the original input data by moving average kernel. DLinear~\cite{Dlinear} uses a simpler linear model to predict time series with more obvious trend and periodicity, and achieves comparable or better performance than various Transformer variants. Recently, TimesNet~\cite{timesnet} proposed a general temporal modeling method on periodic features, which can transform two-dimensional temporal data into three-dimensional, and effectively extract periodic information through visual backbone network. Although these methods take into account trend, periodicity and spatial dependencies, none of them consider the diverse spatial dependencies, which better reflects the spatial information of the real-world.

\section{PRELIMINARIES}
\subsection{Problem Description}
Flood forecasting is based on the historical hydrological data recorded by various monitoring stations in a region, to predict the flow values of that region in a future period of time. We use $X_{t} \in \mathbb{R}^{1 \times N}$ and $X_t^i\in \mathbb{R}^{1\times 1}$ to denote the data from $N$ stations and the $i$-th station at time $t$, respectively. Rain data is represented when $i\ne N$, otherwise it is flow data. Historical data of time length $T$ is denoted by $X=\{X_{t-T+1},X_{t-T+2},\cdots,X_t\}\in \mathbb{R}^{T\times N}$. The predefined spatial structure graph is denoted by $G=\left(V,E,A\right)$, where $V$ is the set of vertices, $E$ is the set of edges, and $A\in \mathbb{R}^{N\times N}$ is the adjacency matrix of $G$.

Given a graph $G$ and historical data $X$ of time length $T$. We need to learn that a function $f$ is used to predict the flow values $Y=\{Y_{t+1},\cdots,Y_{t+H}\}\in \mathbb{R}^H$ for the next $H$ time steps, $Y_{t+1}$ is a scalar indicating the predicted flow value at the $i$-th time step after time $t$. The above content can be expressed by the following mapping relationship:
\begin{equation}
[X_{(t-T+1)},\cdots,X_t;G]\overset{f}{\longrightarrow}[Y_{(t+1)},\cdots,Y_{(t+\mathrm{H})}]
\end{equation}

\subsection{LSTM Method}
LSTM can learn and memorize long-term dependencies in sequences, thus improving the predictive ability of the model. LSTM uses the forget gate $f_t$ to discard information from the previous cell state $c_{t-1}$, the input gate $i_t$ to add information from the candidate cell state ${\widetilde{c}}_t$ to the current cell state $c_t$, and the output gate $o_t$ to output information from the cell state. The computation of the forget gate $f_t$, the input gate $i_t$, and the output gate $o_t$ is shown in Eq.~\ref{eq:lstm-1-end}.

\begin{equation}
    \begin{aligned}
    f_t&=\sigma(W_f\odot[h_{t-1},x_t]+b_f)\\
    i_t&=\sigma(W_i\odot[h_{t-1},x_t]+b_i)\\
    o_t&=\sigma(W_o\odot[h_{t-1},x_t]+b_o) \label{eq:lstm-1-end}
    \end{aligned}
\end{equation}
The current candidate cell state ${\widetilde{c}}_t$, the current cell state ${c}_t$ and the current output ${h}_t$ are shown in Eq.~\ref{eq:lstm-2-end}.

\begin{equation}
    \begin{aligned}
    \tilde c_t&=\tanh(W_c\odot[h_{t-1},x_t]+b_c)\\
    c_t&=f_t\odot c_{t-1}+i_t\odot\tilde c_t\\
    h_t&=o_t\odot\tanh(c_t)  \label{eq:lstm-2-end}
    \end{aligned}
\end{equation}
where $\sigma$ denotes the sigmoid activation function and $\odot$ denotes the hadamard product.

\begin{figure*}[!t]
\begin{center}
\includegraphics[width=\linewidth]{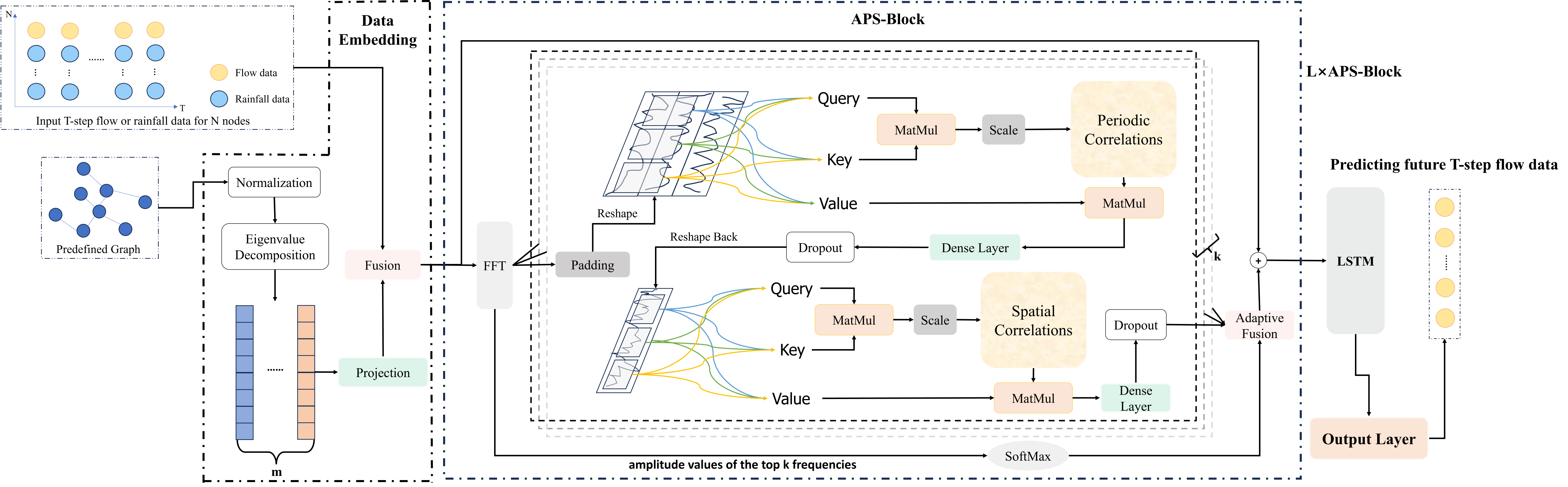}
\end{center}
\caption{The Overall Framework of APS-LSTM.}
\label{fig:model}
\vspace{-0.4cm}
\end{figure*}

\section{METHODOLOGY}
The overall architecture of the APS-LSTM model is shown in Fig.~\ref{fig:model}. The input data and predefined graph structures are first processed by the data embedding layer. In order to capture the deeper hidden periodic and spatial information, we process the embedded data through L layers of the adaptive periodic and spatial self-attention block (APS-Block), which are connected by residual networks~\cite{residual}. Then the data with multi-periodicity and diverse spatial dependencies are fed into LSTM for sequence encoding and ultimately decoded through a linear output layer.

\subsection{Data Embedding}
The main role of the data embedding layer is to process the predefined graph structure using graph Laplacian eigenvectors~\cite{laplacian}, which allows the input data to contain the necessary spatial information.

We first convert the original adjacency matrix to a normalized Laplacian matrix by $L=I-D^{-\frac{1}{2}}AD^{-\frac{1}{2}}$, where $I$ is the identity matrix and $D$ is the degree matrix. The normalized Laplacian matrix is then subjected to an eigenvalue decomposition $L=Q^\top\mathbf{\Lambda}Q$, where $Q$ is the eigenvector matrix and $\mathbf{\Lambda}$ is the eigenvalue matrix. We select the $m$ smallest nontrivial eigenvectors among the eigenvector matrix~\cite{pdformer} and obtain the embedding space $X_s\in \mathbb{R}^N$ by linear projection of these $m$ smallest nontrivial eigenvectors. We fuse the spatial graph Laplacian embedding with the original data using addition of the corresponding elements $X_e=X+X_s\in \mathbb{R}^{T\times N}$.

\subsection{Adaptive Periodic and Spatial Self-Attention Block}
The APS-Blocks are connected by residual networks. Each APS-Block mainly consists of periodic information extraction, periodic and spatial self-attention, and adaptive aggregation.

\subsubsection{Periodic Information Extraction}
Inspired by TimesNet~\cite{timesnet}, we use FFT to convert time-domain data into frequency-domain data. By analyzing frequency and amplitude, we can calculate different periodic division methods and uncover potential multi-periodicity. These can be given by:
\begin{equation}
    \begin{aligned}
        &F=\mathrm{Avg}\left(\mathrm{Amp}(\mathrm{FFT}(X_e))\right)\\
        &\{f_1,\cdots,f_k\}=\underset{f_*\in\{1,\cdots,\frac{L}{2}\}}{\operatorname*{arg\text{Topk}}}(F)\\
        &\{p_1,\cdots,p_k\}=\{\left\lceil\frac T{f_1}\right\rceil,\cdots,\left\lceil\frac T{f_k}\right\rceil\}
    \end{aligned}
    \label{eq:pie}
\end{equation}
where $\text{FFT}(\cdot)$ denotes FFT, $\text{Amp}(\cdot)$ denotes calculating amplitude values, $\text{Avg}(\cdot)$ denotes calculating average values from the $N$ dimension, and finally obtaining the amplitude value of each frequency $F\in \mathbb{R}^T$. We select the $k$ frequencies with the highest amplitude from $F$, where $k$ is a hyperparameter, and the period length ${p_1,\cdots,p_k}$ corresponding to each frequency is calculated by $\{\left[\frac{T}{f_1}\right],\cdots,\left[\frac{T}{f_k}\right]\}$.

We can use the formula Eq.~\ref{eq:23d} to convert a 2D tensor into a 3D tensor. $\mathrm{Padding}(\cdot)$ means padding along the time dimension, $\mathrm{Reshape}_{pn_i,pl_i}(\cdot)$ means reshaping the time dimension of the tensor into the shape of the period length $pl_i$ and the number of periods $pn_i$, that is, the variables within and between periods. The hyperparameter $k$ determines the diversity of the periodic divisions.
\begin{align}
    X_r^i=\mathrm{Reshape}_{pn_i,pl_i}(\mathrm{Padding}(X_{in}))\in \mathbb{R}^{pn_i\times pl_i\times N} \label{eq:23d}\\
    i\in\{1,\cdots,k\} \nonumber
\end{align}

\begin{table*}[!ht]
    \caption{Performance of three metrics for different models on TunXi and ChangHua datasets. The best results are marked in bold.}
    \centering
    
    \renewcommand{\arraystretch}{1.0} % This line changes the height
    \begin{adjustbox}{width=1.0\textwidth}
    {\fontsize{9}{11}\selectfont
    \begin{tabular}{|p{0.1cm}|c|ccc|ccc|ccc|ccc|}
    \hline
        \multicolumn{1}{|c|}{Datasets} & \multicolumn{1}{c|}{Methods} & \multicolumn{3}{c|}{T+1 h} & \multicolumn{3}{c|}{T+3 h} & \multicolumn{3}{c|}{T+6 h} & \multicolumn{3}{c|}{Average} \\ \cline{3-14}
        \multicolumn{1}{|c|}{} & \multicolumn{1}{c|}{} & RMSE & MAE & MAPE & RMSE & MAE & MAPE & RMSE & MAE & MAPE & RMSE & MAE & MAPE \\ \hline 

        \multirow{7}{*}{\rotatebox{90}{TunXi}} 
        & GRU & 20.58 & 13.83 & 8.91\% & 43.65 & 29.18 & 17.79\% & 73.71 & 37.83 & 12.87\% & 46.85 & 26.64 & 12.19\% \\ 
        & LSTM & 24.03 & 19.70 & 18.63\% & 35.31 & 18.78 & 8.12\% & 70.69 & 34.04 & 13.09\% & 43.49 & 23.40 & 11.25\% \\ 
        & DLinear & \textbf{15.31} & \textbf{11.06} & 8.28\% & 40.30 & 19.30 & 9.05\% & 82.17 & 38.27 & 12.92\% & 47.69 & 23.47 & 10.30\% \\ 
        & STA-LSTM & 43.10 & 27.25 & 12.80\% & 51.39 & 30.32 & 12.33\% & 80.65 & 45.59 & 17.08\% & 58.00 & 33.86 & 13.55\% \\ 
        & GraphWaveNet & 24.56 & 16.70 & 9.14\% & 36.67 & 21.95 & 9.06\% & 74.04 & 47.51 & 18.32\% & 45.51 & 28.73 & 11.70\% \\ 
        & TimesNet & 27.87 & 17.90 & 9.32\% & 39.50 & 21.85 & 9.20\% & 69.08 & 34.94 & \textbf{11.30}\% & 45.94 & 25.03 & 9.93\% \\
        & APS-LSTM & 21.58 & 12.63 & \textbf{5.87}\% & \textbf{34.47} & \textbf{17.89} & \textbf{6.68\%} & \textbf{67.66} & \textbf{33.27} & 13.48\% & \textbf{41.70} & \textbf{21.11} & \textbf{8.00\%} \\ \hline

        \multirow{7}{*}{\rotatebox{90}{ChangHua}} 
        & GRU & 27.85 & 14.45 & 30.52\% & 48.06 & 20.43 & 27.12\% & 85.32 & 40.54 & 27.09\% & 55.57 & 25.20 & 24.95\% \\ 
        & LSTM & 30.07 & 14.87 & 15.03\% & 53.6 & 25.36 & 20.67\% & 79.03 & 40.47 & 28.37\% & 55.72 & 26.65 & 22.18\% \\ 
        & DLinear & 27.46 & \textbf{11.34} & \textbf{11.12}\% & 56.66 & 29.05 & 20.92\% & 86.04 & 47.67 & 33.80\% & 59.89 & 31.26 & 22.85\% \\ 
        & STA-LSTM & 32.00 & 16.50 & 14.17\% & 44.90 & 22.31 & \textbf{16.00}\% & 77.15 & 41.01 & 32.11\% & 52.64 & 26.84 & 21.27\% \\ 
        & GraphWaveNet & 30.22 & 15.20 & 14.62\% & 45.83 & 21.53 & 18.26\% & 73.15 & 37.05 & 37.52\% & 50.59 & 24.36 & 21.68\% \\ 
        & TimesNet & 31.28 & 15.11 & 14.87\% & 45.05 & 20.94 & 20.35\% & 79.32 & 37.40 & 25.97\% & 52.76 & 24.97 & 19.58\% \\ 
        & APS-LSTM & \textbf{26.51} & 13.47 & 13.20\% & \textbf{40.44} & \textbf{20.05} & 17.97\% & \textbf{65.44} & \textbf{34.35} & \textbf{25.42}\% & \textbf{44.86} & \textbf{23.00} & \textbf{19.12}\% \\ \hline

    \end{tabular}}
    \end{adjustbox}
\label{tab:result}
\vspace{-0.4cm}
\end{table*}

\subsubsection{Periodic and Spatial Self-Attention}
Attention mechanism~\cite{attention} can effectively focus on the key information and dynamic dependencies in the data. We design Periodic Self-Attention (PSA) and Spatial Self-Attention (SSA) to model the complex, dynamic and diverse periodic and spatial relationships. 

We use Scaled Dot-Product Attention~\cite{transformer} to calculate PSA, as shown in Eq.~\ref{eq:psa1}-\ref{eq:psa2}. The difference is that, because the 2D kernel can pay attention to the change trend within and across periods, Query, Key and Value are calculated by convolution operations, rather than simple linear projection operations. Finally, the attention score of each node in different periods can be calculated.
\begin{equation}
    \begin{aligned}
        Q_{i}^{(p)} &=\mathrm{Conv}2\mathrm{d}_\mathrm{i}^\mathrm{Q}(X_r^i)  \\
        {K_{i}}^{(p)} &=\mathrm{Conv2d_{i}^{K}}(X_{r}^{i})\\
        V_{i}{}^{(p)} &=\mathrm{Conv2d_{i}^{V}(X_{r}^{i})}
    \end{aligned}
    \label{eq:psa1}
\end{equation}
\begin{equation}
    {PSA}_i
    =\text{SoftMax}\left(\frac{{Q_i}^{(p)}{\left({K_i}^{(p)}\right)}^T}{\sqrt{d_{pl_i}}}\right)V_i^{(p)}
    \label{eq:psa2}
\end{equation}
where Conv2d means using 2D kernel for convolution operation.

We transform the data weighted by periodic self-attention back to the 2D tensor $X_p^i\in \mathbb{R}^{T\times N}$ , and then use SSA to learn the spatial dependence under the corresponding period division. Here we utilize 1D kernel to calculate Query, Key and Value, and the other parts are the same as calculating PSA, formulated as follows:
\begin{equation}
    \begin{aligned}
    &Q_{i}^{(s)}=\mathrm{Conv}1\mathrm{d_{i}^{Q}(X_{p}^{i})} \\
    &{K_{i}}^{(s)}=\mathrm{Conv}1\mathrm{d_{i}^{K}}(X_{p}^{i})\\
    &V_{i}^{(s)}=\mathrm{Conv}1\mathrm{d}_{\mathrm{i}}^{\mathrm{V}}(X_{p}^{i})
    \end{aligned}
\end{equation}
\begin{equation}
    \begin{aligned}
    {SSA}_i
    =\text{SoftMax}\left(\frac{{Q_i}^{(s)}({K_i}^{(s)})^T}{\sqrt{d_N}}\right){V_i}^{(s)}
    \end{aligned}
\end{equation}
where Conv1d means using 1D kernel for convolution operation. After weighting by spatial attention score, we transform $X_p^{i}$ into $X_{ps}^i\in \mathbb{R}^{T\times N}$.

\subsubsection{Adaptive Aggregation}
We reshape the data based on different frequencies into $k$ types of 3D tensors, which are transformed by their own periodic and spatial self-attention modules, resulting in $k$ types of outputs, combined as $\{X_{ps}^1,\cdots,X_{ps}^k\}\in \mathbb{R}^{k\times T\times N}$. In order to effectively fuse these $k$ outputs, we use the amplitude values of the top $k$ frequencies calculated before to represent the importance of these $k$ tensors\cite{timesnet,Autoformer}, and then perform weighted summation, calculated as follow: 
\begin{equation}
    \begin{aligned}
        W_{f_1},\cdots,W_{f_2}&=\text{SoftMax}(F_{f_1},\cdots,F_{f_k})\\
        X_{ps}&=\sum\limits_{i=1}^{k} W_{f_i}X_{ps}^{i}\in \mathbb{R}^{T\times N}
    \end{aligned}
    \label{eq:ada}
\end{equation}
The aggregated results encompass features of multiple periods as well as a variety of spatial dependencies.

\section{EXPERIMENTS}
\begin{table}
    \caption{Ablation Study on TunXi and ChangHua Datasets.}
    \centering
    \renewcommand{\arraystretch}{0.55} % This line changes the height
    \begin{adjustbox}{width=\linewidth}
    {\fontsize{5}{11}\selectfont
    \begin{tabular}{|c|c|ccc|}
    \hline
        \multicolumn{1}{|c|}{Datasets} & \multicolumn{1}{|c|}{Methods} & \multicolumn{3}{|c|}{Average} \\ \cline{3-5}
        \multicolumn{1}{|c|}{} & \multicolumn{1}{c|}{} & RMSE & MAE & MAPE \\ \hline 

        \multirow{3}{*}{TunXi} 
        & AP-LSTM & 43.59 & 22.86 & 10.64\% \\
        & AS-LSTM & 42.64 & 22.61 & 11.06\% \\
        & APS-LSTM & \textbf{41.70} & \textbf{21.11} & \textbf{8.00\%} \\ \hline

        \multirow{3}{*}{ChangHua} 
        & AP-LSTM & 49.37 & 24.08 & 20.31\% \\
        & AS-LSTM & 51.07 & 23.29 & \textbf{17.30}\% \\
        & APS-LSTM & \textbf{44.86} & \textbf{23.00} & 19.12\% \\ \hline

    \end{tabular}}
    \end{adjustbox}
\label{tab:albation}
\vspace{-0.6cm}
\end{table}

\subsection{Datasets}
We selected two basins from Chinese regions for the experimental study, where missing values were filled by linear interpolation.\\
\textbf{TunXi Basin}: TunXi located in Tunxi District, Huangshan City, Anhui Province, China, with a watershed area of 2,696.76 km$^2$, including 11 rainfall stations and 1 flow station, containing a total of 43,421 samples.\\
\textbf{ChangHua Basin}: ChangHua located in Lin'an City, Zhejiang Province, China, with a watershed area of 3444 km$^2$ , including 7 rainfall stations and 1 flow station, containing a total of 9370 samples.

\subsection{Baselines}
We compared the APS-LSTM model with six baselines. Two of the baselines are RNN variants: LSTM~\cite{lstm} and GRU~\cite{gru}. Two periodic and trend-based time-series prediction methods: DLinear~\cite{Dlinear} and TimesNet~\cite{timesnet}. Two time-series prediction methods with spatial information: STA-LSTM~\cite{sta-lstm} and Graph WaveNet~\cite{GWnet}.

\subsection{Experiment Settings}
All our experiments were performed on an Ubantu 20.04 system equipped with an NVIDIA 2070Super GPU and 62GB of RAM. Our proposed APS-LSTM model was implemented based on Python 3.10 and PyTorch 1.13. We use the Adam optimizer with 0.01 learning rate for model training and MSE as the loss function. The random seed is set to 2, the batch size is 200, both $L$ and $k$ are both set to 2, and the hidden layer dimension is searched between 80 and 90. The number of training epochs on the TunXi and ChangHua datasets are 60 and 100, respectively.
\begin{equation}
    \begin{aligned}
     RMSE&=\sqrt{\frac{1}{n}\sum\limits_{i=1}^{n}(\hat y_i-y_i)^2}\\
     MAE&=\frac{1}{n}\sum\limits_{i=1}^n|\hat y_i-y_i|\\
     MAPE&=\frac{1}{n}\sum\limits_{i=1}^{n}|\frac{\hat y_i-y_i}{y_i}|
     \label{eq:metrics}
    \end{aligned}
\end{equation}

We use 12 hours of historical data to predict the flood flow in the next 6 hours. The dataset is split into 80\% training set, 5\% validation set and 15\% test set. The model parameters that perform the best on the validation set are used as the final results, and the metrics are evaluated on the test set. We use three evaluation metrics, RMSE, MAE and MAPE, to validate the model performance on the test set, as shown in Eq.~\ref{eq:metrics}. To reduce the impact of smaller values in the sample on MAPE, we mask out the flow values less than 1 when calculating MAPE.
% where $\hat y_i$ and $y_i$ are respectively the predicted values and the true values for the $i$-th sample.

We use Min-Max Normalization to standardize the input data to the range of -1 and 1, as shown in Eq.~\ref{min-max}. To ensure the validity of the experiment, we only calculate the maximum and minimum values in the training set $X_{tr}$ to normalize the data. When evaluating on the test set, we restore the normalized data to the original scale.
\begin{equation}
    X_{norm}=2\frac{X-\min(X_{tr})}{\max({X_{tr}})-\min({X_{tr}})}-1
    \label{min-max}
\end{equation}

\begin{figure}[!t]
    \centering
    \begin{minipage}{\linewidth}
        \centering
        \includegraphics[width=\linewidth]{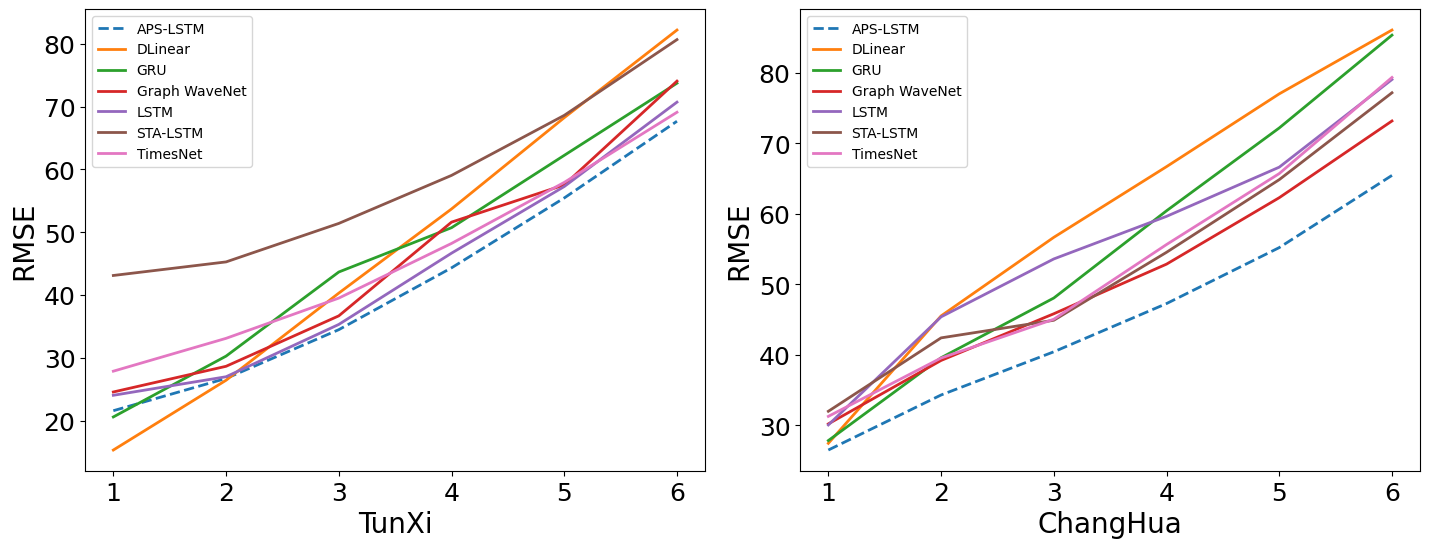}
        \caption{Trends in RMSE from T+1 to T+6 for different models}
        \label{fig:RMSE}
    \end{minipage}
    
    \vspace{5pt} % Adjust vertical spacing between figures

    \begin{minipage}{\linewidth}
        \centering
        \includegraphics[width=\linewidth]{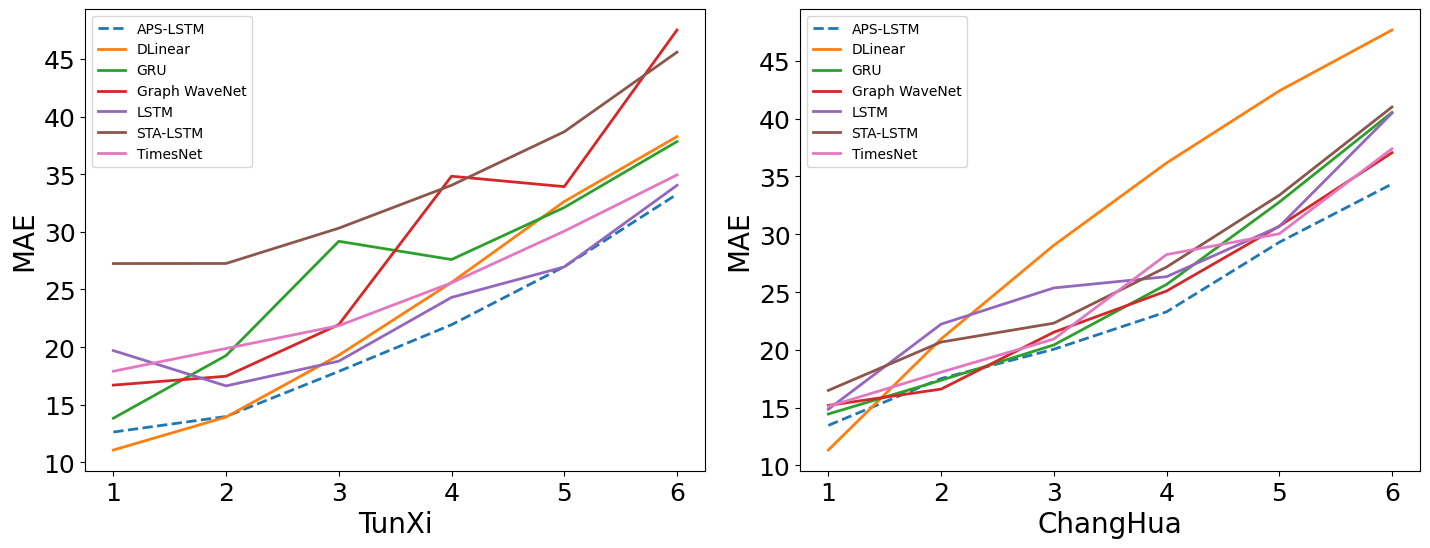}
        \caption{Trends in MAE from T+1 to T+6 for different models}
        \label{fig:MAE}
    \end{minipage}
    
    \vspace{5pt} % Adjust vertical spacing between figures

    \begin{minipage}{\linewidth}
        \centering
        \includegraphics[width=\linewidth]{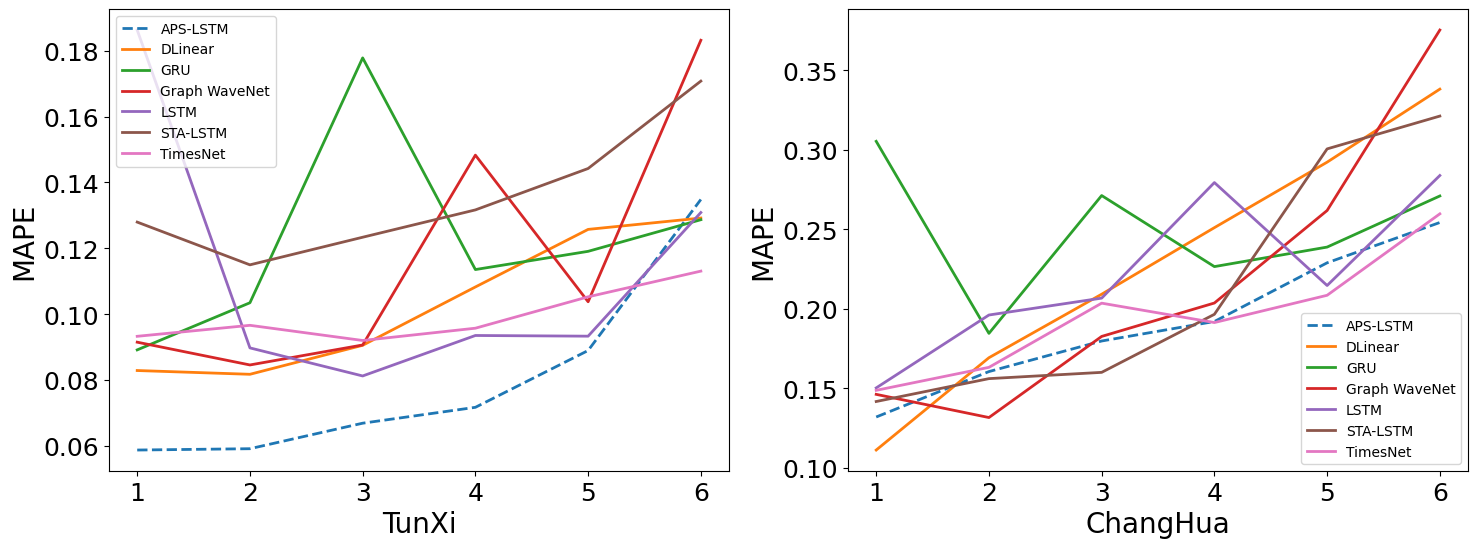}
        \caption{Trends in MAPE from T+1 to T+6 for different models}
        \label{fig:MAPE}
    \end{minipage}
\vspace{-0.6cm}
\end{figure}

\subsection{Analysis and Comparison of Results}
The models use 12 hours of historical data on the TunXi and ChangHua datasets to predict the next 6 hours in the flood flow data. Table~\ref{tab:result} shows the evaluation metrics for the future 1st hour (T+1), 3rd hour (T+3), 6th hour (T+6), and the 6-hours average. Our proposed APS-LSTM performs best on average compared to the other 6 baselines. In terms of RMSE and MAE, APS-LSTM achieves the best results except for T+1. In terms of MAPE metrics, it also maintains a high level of best or second-best performance at almost all times. On the TunXi dataset, the average RMSE, MAE, and MAPE metrics of APS-LSTM are reduced by 4.1\%, 9.8\%, and 19.4\%, respectively, compared to the second-best model. On the ChangHua dataset, the reductions are 11.3\%, 5.6\% and 2.3\%, respectively.

We also show the complete change trends of the evaluation metrics for APS-LSTM and 6 baselines when predicting the future 6 hours of flood flow. As shown in Fig.~\ref{fig:RMSE} and Fig.~\ref{fig:MAE}, as the prediction time length increases, the RMSE and MAE of all models show an upward trend. But APS-LSTM performs better, with a slower upward trend on these two metrics, and leads all the baselines from T+3 onwards. Fig.~\ref{fig:MAPE} shows the MAPE of each model in the experiment, the increase of time span clearly causes some fluctuations in the MAPE of 6 baselines. This is caused by the over-prediction of smaller values, which has little impact on the practical application of the flood prediction task. Compared with other methods, our proposed APS-LSTM is more stable, thus achieving the best performance with the smallest average MAPE.

To assess the effectiveness of the periodic self-attention and spatial self-attention modules, we disable the spatial self-attention and periodic self-attention components respectively, resulting in two weakened versions: AP-LSTM and AS-LSTM. As can be seen from Table~\ref{tab:albation}, APS-LSTM is only slightly inferior to AS-LSTM in terms of MAPE metrics for the ChangHua dataset, but achieves absolute advantages in other aspects, with a more comprehensive performance. 
% This is the superior performance achieved by the combined effect of periodic self-attention and spatial self-attention.

\begin{figure}[!t]
    \centering
    \begin{minipage}{0.49\linewidth}
        \centering
        \includegraphics[width=\linewidth]{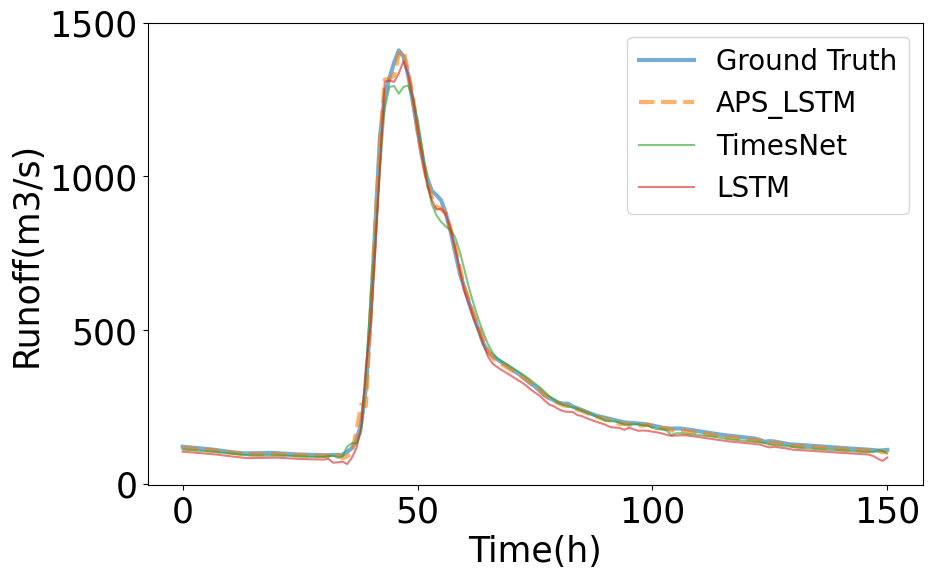} 
        \caption{Predicting at T+1}
        \label{fig:T+1}
    \end{minipage}
    \begin{minipage}{0.49\linewidth}
        \centering
        \includegraphics[width=\linewidth]{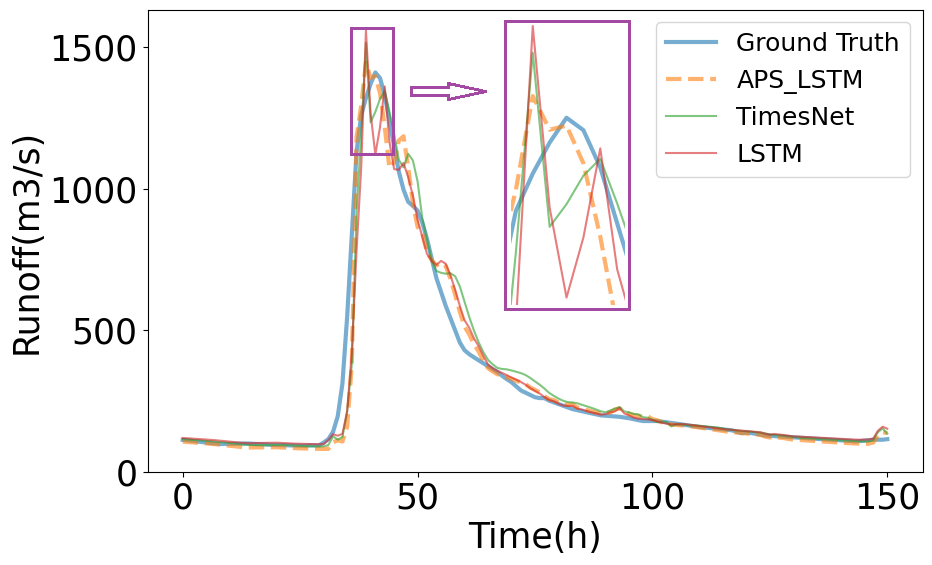} 
        \caption{Predicting at T+6}
        \label{fig:T+6}
    \end{minipage}
\vspace{-0.6cm}
\end{figure}

We select a flood scene in the TunXi test set, and show the peak flow prediction performance of the three best-performing models in this flood event at moments T+1 and T+6, as shown in Fig.~\ref{fig:T+1} and Fig.~\ref{fig:T+6}. The prediction trends of APS-LSTM, LSTM and TimesNet are similar in most cases, but they differ when predicting the flood peak. APS-LSTM can capture the multi-period and diverse spatial dependencies better than LSTM and TimesNet, so it can adjust the prediction direction and make the peak flow prediction more accurate. At T+1, APS-LSTM predicted the peak time and peak flow almost in line with the actual situation, performing the best. LSTM showed a significant lag, while TimesNet predicted the peak flow as a V-shaped trend, performing the worst. At T+6, although the peak times predicted by the three models are earlier than the actual time, APS-LSTM predicted the flow value closer to the actual value and did not show a noticeable V-shaped change trend like LSTM and TimesNet.
% APS-LSTM’s prediction results fitted the actual peak flow change trend better.

\subsection{Visualization of Periodic and Spatial Correlations}
\begin{figure}[!t]
    \centering
    \begin{minipage}{\linewidth}
        \centering
        \includegraphics[width=\linewidth]{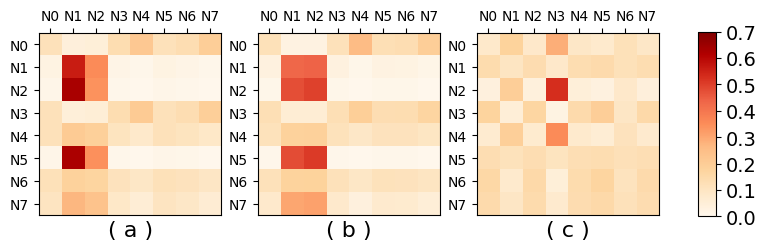} 
        \caption{Case Study of Spatial Self-Attention}
        \label{fig:spatial}
    \end{minipage}
    
    \vspace{2pt}

    \begin{minipage}{\linewidth}
        \centering
        \includegraphics[width=\linewidth]{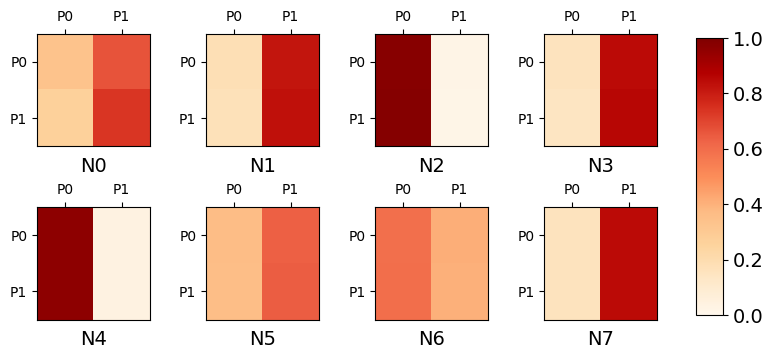} 
        \caption{Case Study of Periodic Self-Attention}
        \label{fig:periodic}
    \end{minipage}
\vspace{-0.6cm}
\end{figure}

To increase the interpretability of the model, we selected a sample from ChangHua test set, and then visualized the spatial correlations between the nodes in that sample, as shown in Fig.~\ref{fig:spatial}. First, Fig.~\ref{fig:spatial}(a) and Fig.~\ref{fig:spatial}(b) are compared, which show the changes in the spatial association degree between the stations in different time periods under the same period divisions. Then, we compared Fig.~\ref{fig:spatial}(a) and Fig.~\ref{fig:spatial}(c), which illustrate the shifts in the spatial attention focus between the stations in the same time period under different period divisions. The qualitative analysis above indicates that our designed spatial self-attention effectively captures the dynamic spatial structures under different periodic divisions. Fig.~\ref{fig:periodic} depicts the periodic self-attention weights, and the eight sub-figures correspond to the periodic correlations of the eight stations respectively. N1, N2, N3, N4, and N7 show a strong focus of attention on a particular side of the periods, while the other three stations are more balanced. The various periodic self-attention weights enable APS-LSTM to capture rich periodic characteristics.

\section{CONCLUSIONS}
In this paper, we propose a novel model for flood flow prediction, named APS-LSTM. Our model uncovers elusive periodic characteristics and captures diverse spatial dependencies in hydrological data. Through quantitative and qualitative analyses on two hydrological datasets, we show that APS-LSTM outperforms six existing baselines in flood flow prediction. Our findings highlight the importance of mining multi-periodicity and diverse spatial dependencies for accurate flood prediction. In future work, we aim to integrate these aspects with other architectures to further improve flood forecasting accuracy.

\bibliographystyle{IEEEtran}
\bibliography{main}

\end{document}